\documentclass[letterpaper, 10 pt, journal, twoside]{IEEEtran}  %

\usepackage{cite}
\usepackage[hidelinks]{hyperref}
\usepackage{amsmath,amssymb,amsfonts}
\usepackage{graphicx}
\usepackage{xcolor}
\usepackage{tabularx}
\usepackage{multirow}
\usepackage{algorithm}
\usepackage[noend]{algpseudocode}
\usepackage{adjustbox}
\usepackage{makecell}
\usepackage{booktabs} 
\usepackage[singlelinecheck=false]{caption}
\usepackage{subcaption}
\usepackage{lettrine}
\usepackage[deletedmarkup=sout, defaultcolor=orange]{changes} 

\ifCLASSINFOpdf
\else
\fi

\begin{document}
\title{Traversability-aware Adaptive Optimization for Path Planning and Control in Mountainous Terrain}

\author{Se-Wook Yoo$^{1*}$, E-In Son$^{1*}$, and Seung-Woo Seo$^{1}$
    \thanks{Manuscript received: February, 2, 2024; Accepted March, 23, 2024.}
    \thanks{This paper was recommended for publication by Editor Hanna Kurniawati upon evaluation of the Associate Editor and Reviewers' comments. This work was supported by the Automation and Systems Research Institute (ASRI), and the BK21 FOUR program of the Education and Research Program for Future ICT Pioneers, Seoul National University in 2024.}
    \thanks{$^{*}$Equal contribution. $^{1}$The authors are with Department of Electrical and Computer Engineering, Seoul National University, Seoul, Republic of Korea.
        {\tt\footnotesize \{tpdnr1360, pingpang, sseo\}@snu.ac.kr}}
    \thanks{Digital Object Identifier (DOI): see top of this page.}
}

\markboth{IEEE Robotics and Automation Letters. Preprint Version. Accepted March, 2024}
{Yoo and Son \MakeLowercase{\textit{et al.}}: Traversability-aware Adaptive Optimization for Path Planning and Control in Mountainous Terrain} 

\maketitle

\begin{abstract}

Autonomous navigation in extreme mountainous terrains poses challenges due to the presence of mobility-stressing elements and undulating surfaces, making it particularly difficult compared to conventional off-road driving scenarios. In such environments, estimating traversability solely based on exteroceptive sensors often leads to the inability to reach the goal due to a high prevalence of non-traversable areas. In this paper, we consider traversability as a relative value that integrates the robot's internal state, such as speed and torque to exhibit resilient behavior to reach its goal successfully. We separate traversability into apparent traversability and relative traversability, then incorporate these distinctions in the optimization process of sampling-based planning and motion predictive control. Our method enables the robots to execute the desired behaviors more accurately while avoiding hazardous regions and getting stuck. Experiments conducted on simulation with 27 diverse types of mountainous terrain and real-world demonstrate the robustness of the proposed framework, with increasingly better performance observed in more complex environments.
\end{abstract}


\IEEEpeerreviewmaketitle

\section{Introduction}

\IEEEPARstart{A}{s} research on the autonomous navigation of ground robots in off-road environments is actively progressing, its utilization is gradually increasing in the fields of defense, agriculture, and industry \cite{papadakis2013terrain, santos2021navigation, zhao2020new}. Compared with urban environments, off-road environments have various obstacles that hinder movements, such as bushes and boulders, and the geometry is complicated due to the unstructured and uneven terrain \cite{hines2021virtual}. In particular, mountainous terrain poses a greater challenge due to its extreme variance in geometrical features such as steep slopes, high elevations, and rugged terrain \cite{paton2020navigation}.

Early study \cite{laubach1998autonomous} applies binary classification to terrain in order to classify it as either traversable or non-traversable for navigation on uneven terrain. \cite{castelnovi2005reactive} broadens the scope of the terrain classification from binary to multi-class and assigns a fixed speed profile for each class. However, discrete categorization is insufficient to naturally capture the terrain feature. To overcome this problem, \cite{pivtoraiko2009differentially} maps the terrain properties produced by perceptual data to a scalar cost value, so-called traversability. Such approaches consider designed traversability with hand-crafted metrics such as slope, slippage, and flatness in the optimization process. \cite{jian2022putn} uses exteroceptive sensor-based traversability as the coefficient of the control cost to slow down the robot in dangerous areas. However, in mountainous terrain where the overall traversability cost is high, it leads to frequent instances of the robot getting stuck due to the reduced velocity. To address this issue, it is necessary to consider the traversability cost in a relative perspective according to the robot's motion, rather than solely relying on geometric representation. For instance, a low velocity may prevent the robot from moving through actually traversable areas, while a high velocity may allow the robot to move through untraversable areas. Therefore, we incorporate proprioceptive sensor-based traversability, which takes into account the inertial state of the robot, as the tracking coefficient to adjust the robot's velocity dynamically. In addition, to prevent our method from overlooking the potential risk in hazardous terrain, we adopt constraints in optimization to limit the robot's behavior.

\begin{figure}    
    \centering
    \includegraphics[width=0.87\columnwidth]{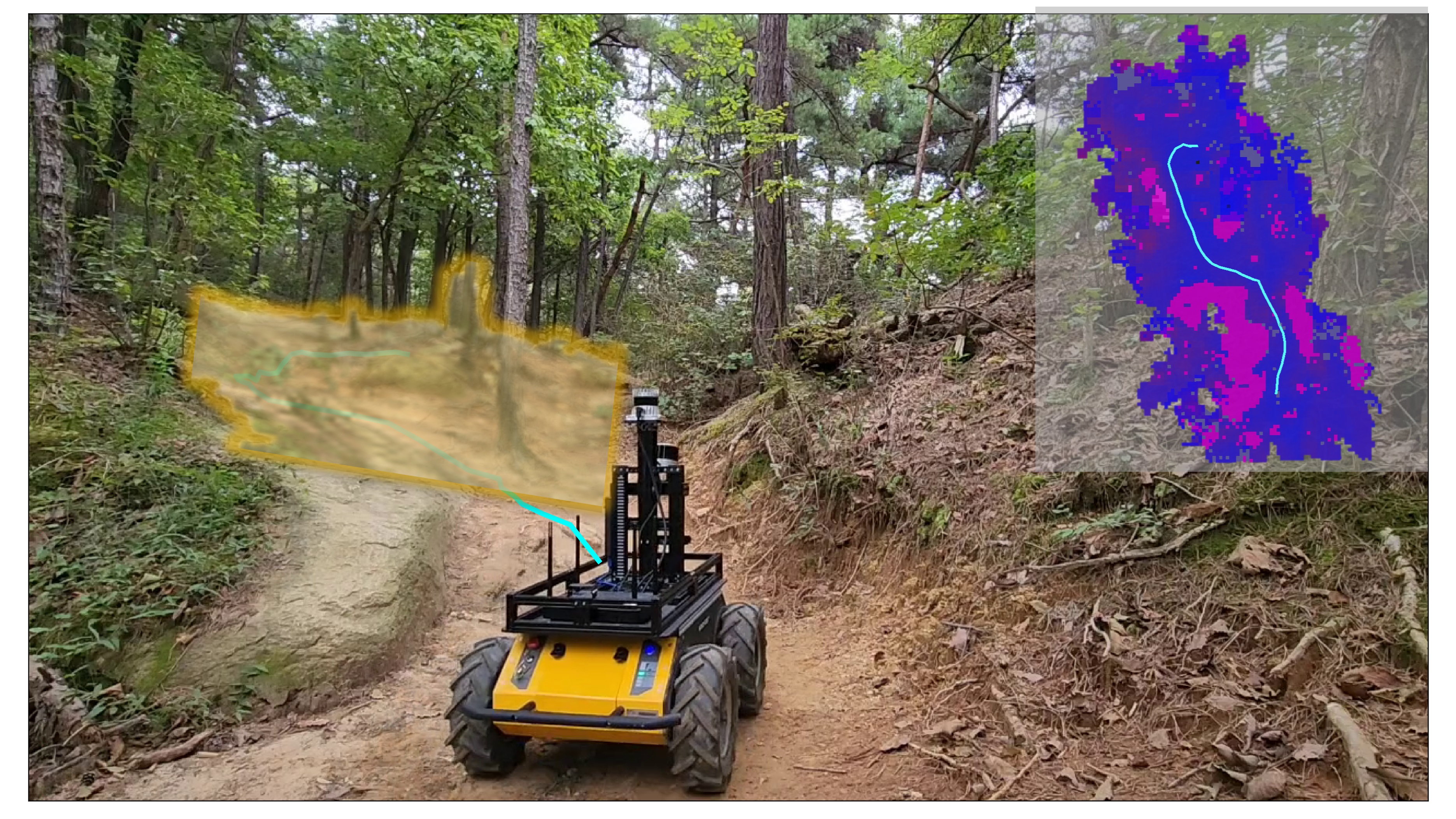}
    \caption{\textbf{Robot Navigation in Mountainous Terrain.} We propose a traversability-aware navigation framework in both virtual and real environments. Virtual environments are reconstructed from actual mountainous terrain. The top right of the figure displays the optimal path (cyan) calculated using traversability information.}
    \vspace{-0.4cm}
    \label{fig1:target_domain}  
\end{figure}

\begin{figure*}
    \centering
    \includegraphics[width=0.85\linewidth]{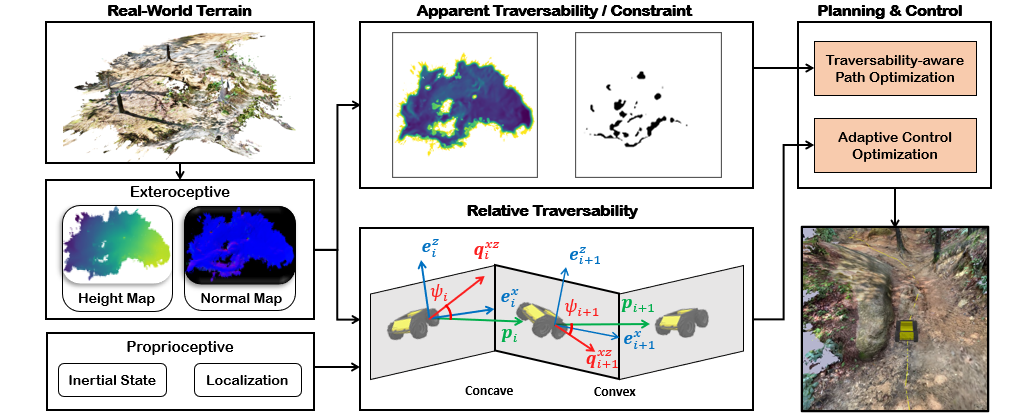}
    \vspace{-0.1cm}
    \caption{\textbf{Overall Structure.} Our framework takes inputs such as a height/normal map, localization, and the robot's inertial states (orientation $\mathbf{q}$, linear velocity $\mathbf{\zeta}$, and angular velocity $\mathbf{\omega}$). The computed traversability/constraint map of the terrain is used as guidance to generate a trajectory. The controller adjusts the robot's velocity with adaptive optimization based on apparent and relative traversability. This enables the robot to navigate extreme mountainous terrains robustly. The top left of the figure shows a sample of terrains collected on campus and mountains.}
    \vspace{-0.4cm}
    \label{fig2:overall_structure}    
\end{figure*}

In this paper, we present \textbf{TAO}, \textbf{T}raversability-aware \textbf{A}daptive \textbf{O}ptimization in mountainous terrain to assist in flexible decision-making processes for robot navigation. We design traversability into two components: apparent traversability using exteroceptive features and relative traversability using proprioceptive features, which consider the ground-robot interaction through the robot's posture variability. Subsequently, we integrate the traversability into the optimization process of sampling-based planning and motion predictive control. Apparent traversability is used as a sampling bias for planning to generate a feasible path. This effectively reduces the search space while generating a feasible path that circumvents undulating regions, eliminating the high potential risk. Relative traversability is applied to regulate the robot's velocity to execute the desired behaviors more accurately while avoiding vigorous posture change, getting stuck, or being damaged.

The main contributions can be summarized as follows:

\begin{itemize}
\item We introduce a novel adaptive navigation method that takes into account the relative perspective of traversability based on the robot's inertial state.

\item We achieve an improvement of up to 16\% in both effectiveness and stability through hazard avoidance and dynamic velocity adjustments that take into account ground-robot interaction.

\item We demonstrate the feasibility of our method to various types of off-road terrain through experiments conducted in real-world and simulations involving 27 diverse terrains.

\end{itemize}

The remainder of this paper is organized as follows. A comparative review of related works is presented in Section \ref{related}. We state the problem definition in Section \ref{problem}. The proposed framework is explained in detail in Section \ref{method}. Experimental results are demonstrated in Section \ref{experiment}. The conclusion follows in Section \ref{conclusion}.
\vspace{+0.2cm}

\section{Related Work} \label{related}

In this section, we discuss existing work on modeling terrain traversability and acquiring resilient skills for robot navigation in off-road environments.

\subsection{Modelling Terrain Traversability}

The utilization of exteroceptive and proprioceptive sensory data is widely studied for modeling terrain traversability. Using exteroceptive sensors consists of appearance-based and geometry-based approaches \cite{papadakis2013terrain}. Appearance-based approaches devise terrain traversability by classifying terrain types according to a category (grass, road, tree, etc.) from visual measurements \cite{gan2020bayesian}. Geometry-based approaches compute the traversability by considering geometrical representations such as slope or roughness from depth measurements \cite{shan2018bayesian}. Unlike using exteroceptive sensors, which ignore the robot's inertial states, proprioceptive approaches model traversability using measurements that capture the robot's inertial state, such as ground reaction score \cite{wellhausen2019should}, vibration \cite{bekhti2020regressed}, or energy consumption \cite{gan2022energy}.

\subsection{Traversability-aware Navigation}

Applying a classic path planner \cite{hart1968formal, koenig2002d, karaman2011sampling} to an uneven off-road terrain poses a challenge since the robot operates in a dynamic environment while avoiding mobility-stressing elements. Initial work assigns a fixed speed profile according to the roughness \cite{castelnovi2005reactive} or semantic units \cite{ojeda2006terrain} of the terrain surface. \cite{tanaka2015fuzzy} estimates the cost map through terrain traversability assessment (TTA) using features on digital elevation map (DEM) or cumulative fractional area (CFA) and utilizes it as a guide of course direction. Specifically, TTA can be added to the distance cost when generating a search graph  \cite{ono2015risk}, or adjusting an existing route \cite{thoresen2021path}. Recently, learning-based methods have been suggested to predict the cost map using data labeled with reachability \cite{daftry2022mlnav, siva2024self, siva2021enhancing}. In \cite{wellhausen2021rough}, the predicted cost map is utilized as adaptive sampling distribution to create a roadmap. Meanwhile, in \cite{weerakoon2022terp}, the cost map is learned using ground interaction data collected by the agent in the reinforcement learning process and used as a bias for course direction.

A large amount of training data is required to generalize a traversable map using a learning-based method. In particular, ground interaction data is scarce or difficult to obtain in off-road terrain, and the labeling cost is expensive. Furthermore, domain adaptation is required to apply the robot to the real-world environment even if the model is trained with simulation data. On the other hand, \cite{jian2022putn} applies rule-based TTA to the cost of the RRT node for path optimization and to find reactive parameters for MPC optimization. However, as inconsistency occurs between desired and actual behavior in tracking the planned path, the cumulative error gradually increases, resulting in a stuck, wheel slip, or tip-over.

Since the mountainous environment is difficult to define all types in advance, our method computes geometry-based TTA using depth measurement without classifying the terrain type.  We also utilize the approach from \cite{jian2022putn} to seamlessly integrate the geometry-based TTA into the path planning and control process. However, we extend this approach by incorporating a traversability distribution with additional consideration of the geometry-based constraint map to apply an adaptive sampling scheme, which is inspired from \cite{wellhausen2021rough}. Furthermore, we propose reactive parameters for optimizing control variables by using the ground reaction score, which considers the change in pitching angle that occurs when the robot interacts with the undulating terrain.

\section{Problem Definition} \label{problem}

Our goal is to develop a framework that enables a robot to navigate through mountainous terrain while ensuring safe and efficient traversal. We define a search space $X \subset \mathbb{R}^{3}$ to achieve this goal. Let us denote the location in search space as $\mathbf{x}=(x,y,z) \in X$, the space occupied by non-ground obstacles as $X_{obs} \subset X $, the area of the ground surface as $X_{surf} \subset X$, and the area of traversable space as $X_{\tau} = X_{surf} \cap X \setminus X_{obs}$. To distinguish traversable area in $X_{\tau}$, we define apparent traversability as $\mathcal{M}_{\tau}$ and regional constraints as $\mathcal{M_{\gamma}}$, respectively, which are expressed by occupancy grid map $\mathcal{M}$. To calculate $\mathcal{M}_{\tau}$, we utilize height map $\mathcal{M}_{z}$ and normal vector map $\mathcal{M}_{\mathbf{n}}$. We also define relative traversability $\Psi_{\tau}$ to consider ground-robot interaction. $\Psi_{\tau}$ represents the fluctuation of the pitching angle that occurs when a robot traverses uneven terrain.

The formal definition of our problem is as follows: path planner searches path $\mathcal{P} \subset X_{\tau}$ from  $\mathbf{x}_{start}$ to $ \mathbf{x}_{goal}$. The optimal path $\mathcal{P}^{*} \subset \mathcal{P}$ is obtained by minimizing $\mathcal{M}_{\tau}$ such that $\mathcal{M}_{\gamma}$ while satisfying to avoid collision with obstacles. The path tracker searches for a feasible control strategy $\pi^{*}$ that minimizes the difference between the desired state and predicted state for the planning time horizon. We also regularize optimization parameters with $\mathcal{M}_{\tau}$ and $\Psi_{\tau}$ to follow $\mathcal{P}^{*}$ reactively considering the interaction of the terrain and the robot.

\section{Methodology} \label{method}

We present TAO, and Fig. \ref{fig2:overall_structure} shows the overall structure. We compute apparent traversability to reflect the geometric features of the surrounding terrain using depth measurements from exteroceptive sensors. To consider ground-robot interaction, we calculate relative traversability using measurements from exteroceptive and proprioceptive sensors. Then, we use apparent traversability as sampling bias and the cost for generating an optimal trajectory. Relative traversability is utilized to dynamically adjust the robot’s velocity, ensuring the tracking controller's ability to follow the given path.

\subsection{Terrain Assessment for Non-planar Ground}

We consider the geometric features of the non-planar ground to traverse extreme terrain. The geometric features comprise a height map $\mathcal{M}_{z}$ and a normal map $\mathcal{M}_{\mathbf{n}}$. To get $\mathcal{M}_{z}$ and $\mathcal{M}_{\mathbf{n}}$, an observed plane is represented as $N$ sets of points included in a square kernel with a side length $K$ centered on $x_i$. A set of $N$ kernels is represented as $\mathcal{K}(\mathbf{x})\ = \{ \mathcal{K}(\mathbf{x}_{i}) \;\vert\; \mbox{for }i = 1, \dots, N \}$. We utilize the SVD algorithm to fit a surface plane $\mathcal{K}(\bar{\mathbf{x}})$ with the set of points $\mathcal{K}(\mathbf{x})$. Here, we denote each point on the fitted plane as $\bar{\mathbf{x}}_{i} = (x_i, y_i, z_i)$ and the corresponding unit normal vector as $\mathbf{n}_i=(n^{x}_{i}, n^{y}_{i}, n^{z}_{i})$. To obtain an occupancy grid representation, we map each point in the kernel to a corresponding point on $\mathcal{M}$ with a size of $\mathbb{R}^{H \times W}$.
\newline 

\begin{figure}
    \vspace{+0.1cm}
    \centering
    \begin{subfigure}[b]{0.2\textwidth}
        \includegraphics[width=\textwidth]{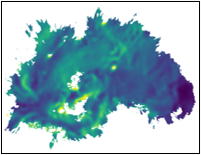}
        \vspace{-0.6cm}
        \captionsetup{justification=centering}
        \caption{Slope}
        \label{fig3a:slope}
    \end{subfigure}
    \begin{subfigure}[b]{0.2\textwidth}
        \includegraphics[width=\textwidth]{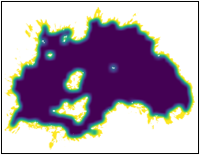}
        \vspace{-0.6cm}
        \captionsetup{justification=centering}
        \caption{Sparsity}
        \label{fig3b:sparsity}
    \end{subfigure}
    \begin{subfigure}[b]{0.2\textwidth}
        \includegraphics[width=\textwidth]{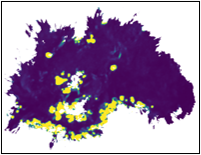}
        \vspace{-0.6cm}
        \captionsetup{justification=centering}
        \caption{Bumpiness}
        \label{fig3c:bumpiness}
    \end{subfigure}
    \begin{subfigure}[b]{0.2\textwidth}
        \includegraphics[width=\textwidth]{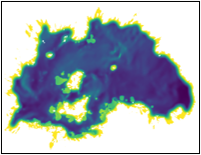}
        \vspace{-0.6cm}
        \captionsetup{justification=centering}
        \caption{Traversability}
        \label{fig3d:traversability}
    \end{subfigure}
    \caption{\textbf{Visualization of Exteroceptive Features.} The color of each map is represented by a range from deep blue with a score of zero to vibrant yellow with a score of one. (d) is obtained as a linear combination of the components (a-c).}    
    \label{fig3:Estimated_Traversability}
    \vspace{-0.3cm}
\end{figure}

\subsubsection{Apparent Traversability and Regional Constraints from Exteroceptive Features}

We utilize the apparent traversability $\mathcal{M}_{\tau}$ as a criterion to select an optimal path with minimal slope and fluctuation. Inspired from \cite{jian2022putn}, the traversability $\tau_{i} \in [0,1]$ for $i$-th kernel is evaluated based on the linear combination of the following exteroceptive features: slope, sparsity, and bumpiness (see Fig. \ref{fig3:Estimated_Traversability})

\vspace{-0.2cm}
\begin{equation}
    \tau_{i} = {\alpha_1} \tilde{\eta}_{slop} + {\alpha_2} \tilde{\eta}_{spar} + {\alpha_3}\tilde{\eta}_{bump},
    \label{eq1:traversabiliy}
\end{equation}

where the weights $\alpha_1$, $\alpha_2$, and $\alpha_3$ sum to 1. $\tau_i = 0$ indicates that the ground in $i$-th kernel is fully traversable, while $\tau_i = 1$ indicates that the ground is non-traversable. $\tilde{\eta}$ represents a normalized value for each feature. Each feature is calculated as follows:

\vspace{-0.2cm}
\begin{align}
    \eta_{slop} &= \arccos (\mathbf{n}_i^{\top} \cdot \mathbf{k})
    \label{eq2:slope} \\
    \eta_{spar} &= \begin{cases}
                    \qquad 1 & r_{i} > r_{max} \\
                    \frac{r_{i} - r_{min}}{r_{max} - r_{min}} & r_{i} \in [r_{min}, r_{max}] \\
                    \qquad 0 & r_{i} < r_{min}
                    \end{cases}
    \label{eq3:sparsity} \\
    \eta_{bump} &= \inf_{\gamma \in \Pi(P_{\mathcal{K}_{i}}, Q_{\bar{\mathcal{K}}})} \mathbb{E}_{x, y \sim \gamma} [\| x - y \|],
    \label{eq4:bumpiness}
\end{align}

where $\mathbf{k}$ is a unit vector in 3-dimensional Euclidean space that is perpendicular to the $xy$ plane. Vacancy ratio $r_{i}$  represents the proportion of vacant parts of $\mathcal{K}(\mathbf{x}_{i})$. $r_{i}$ is determined by $r_{min}$ and $r_{max}$, representing the minimum and maximum values, respectively. If there are clusters of vacant points, it suggests a high possibility of pits and scarps on the ground. Unlike \cite{jian2022putn}, we also incorporate the bumpiness feature, as the slope alone may not fully account for the risk posed by significant terrain fluctuations. We use the earth mover's distance (EMD) \cite{levina2001earth} to compute the bumpiness of the ground. EMD is obtained by comparing the distribution of $\mathcal{K}(\mathbf{x}_{i})$ with the distribution of $\mathcal{K}(\bar{\mathbf{x}})$. Each distribution is denoted as $P_{\mathcal{K}_{i}}$ and $Q_{\bar{\mathcal{K}}}$, respectively. The set $\Pi(P_{\mathcal{K}_{i}}, Q_{\bar{\mathcal{K}}})$ represents all joint distributions that have $P_{\mathcal{K}_{i}}$ and $Q_{\bar{\mathcal{K}}}$ as their respective marginal distributions. Furthermore, we utilize the regional constraints $\mathcal{M}_{\gamma}$ to avoid hazardous terrain on non-planar ground. Based on designed features in Eq. \ref{eq2:slope} - \ref{eq4:bumpiness}, the constraints $\gamma_{i} \in \{0,1\}$ for $i$-kernel is evaluated as follows:

\vspace{-0.3cm}
\begin{equation}
    \resizebox{0.88\columnwidth}{!}{$
        \gamma_{i} = ((\tilde{\eta}_{slop} > \epsilon_{slop}) \lor
        (\tilde{\eta}_{bump} > \epsilon_{bump})) \land
        (\tilde{\eta}_{spar} < \epsilon_{spar}),
    $}
    \label{eq5:regional_constraint}
\end{equation}

where $\epsilon$ is the threshold value for each feature. $\gamma_{i}=1$ indicates that the perceived point on the ground is hazardous.
\newline

\subsubsection{Relative Traversability from Exteroceptive and Proprioceptive Features} 

To robustly follow a given path in undulating terrain, it is essential to consider ground-robot interaction in areas around the robot. Addressing this, we propose relative traversability $\Psi_{\tau}$ that considers the relations among the robot's orientation $\mathbf{q}$, the ground normal vector $\mathbf{n}$, and the planned path direction $\mathbf{p}$. Here, we denote the relative traversability as $\psi_{i} \in \Psi_{\tau}$, where $i$ is the index corresponding to the kernel that the robot is located in $k$-th step of the planned path. As shown in Fig. \ref{fig2:overall_structure}, $\psi_{i}$ is calculated as follows: 

\vspace{-0.1cm}
\begin{equation}
        \mathbf{e}_i^z = \mathbf{n}_i
\end{equation}
\vspace{-0.3cm}
\begin{equation}
        \mathbf{e}_i^x = \frac{\mathbf{p}_i - k_i \mathbf{e}_i^z}{\|\mathbf{p}_i - k_i \mathbf{e}_i^z\|^2}
\end{equation}
\vspace{-0.2cm}
\begin{equation}
       \mathbf{e}_i^y = \mathbf{e}_i^x \times \mathbf{e}_i^z 
\end{equation}
\vspace{-0.3cm}
\begin{equation}
        \mathbf{q}_{i}^{xz} = \mathrm{Proj}_{\mathbf{e}_i^y} (\mathbf{q}_{i}) 
\end{equation}
\vspace{-0.3cm}
\begin{equation}
        \psi_{i} = \arcsin{((\mathbf{q}_{i}^{xz}\times \mathbf{p}_i)^{\top} \cdot \mathbf{e}_{i}^{y})},
\end{equation}
\vspace{-0.1cm}

where $\mathbf{p}_i = {p}_{i+1} - {p}_i$ is directional vector towards the next planned point and $k_i = \mathbf{p}_i^{\top} \cdot \mathbf{e}_i^z$ is the component of the vector $\mathbf{p}_i$ in the direction of $\mathbf{e}_z$. Note that planned points are computed using Eq. \ref{eq13:motion_model}, which considers the robot's linear and angular velocity.

\subsection{Planning and Control}
\begin{figure}[!t]
    \centering
    \vspace{-0.2cm}
    \begin{minipage}{0.9\columnwidth}
        \begin{algorithm}[H]  
        \caption{RejectionSampling($B,\mathcal{M}_{\mathbf{n}}, \mathcal{M}_{z}$)}
            \label{algo1:rejection_sampling}
            \begin{algorithmic}[1]
                \State $\mathcal{N}_{i}$ $\gets$ \textrm{UniformSampling}($B$)
        
                \State Calculate $\tau_{i}$ from Eq. \ref{eq1:traversabiliy} using $\mathcal{M}_{\mathbf{n}}$ and $\mathcal{M}_{z}$
        
                \State Calculate $\gamma_{i}$ from Eq. \ref{eq5:regional_constraint} using $\mathcal{M}_{\mathbf{n}}$ and $\mathcal{M}_{z}$
                
                \State Calculate $\rho_{i}$ from Eq. \ref{eq11:target_distribution} using $\tau_{i}$ and $\gamma_{i}$
        
                \State Sample $\nu$ $\sim$ U(0,1)
        
                \If{$\nu < \rho_{i}$}
                
                    \Return $\mathcal{N}_{i}$
                    
                \Else
                
                    \Return \textrm{RejectionSampling}($B,\mathcal{M}_{\mathbf{n}}, \mathcal{M}_{z}$)
                    
                \EndIf
            \end{algorithmic}
        \end{algorithm}
    \end{minipage}
    \vspace{-0.5cm}
\end{figure}

\subsubsection{Traversability-aware Path Optimization}

We propose an adaptive sampling scheme and path optimization method that can be integrated into a sampling-based search tree method for path planning. This is compatible with various RRT variants and can improve the solution quality. In general, each element of the search tree $\mathcal{T}$ is represented as a node $\mathcal{N}_{i}$ instead of a location $\mathbf{x}_{i}$. We expand this representation as a tuple $\mathcal{N}_{i} = (\mathbf{x}_{i}, \tau_{i}, \gamma_{i})$. The added $\tau_{i}$ and $\gamma_{i}$ are utilized as criteria for tree expansion and path selection. Based on this, we generate a path that circumvents terrains difficult for traversal, instead of just seeking the shortest path. To achieve this, we define a sampling weight $\rho_{i}$ that can provide guidance for sampling new nodes for tree expansion as follows:

\begin{equation}
    \rho_{i} = (1 - \tau_{i})(1 - \gamma_{i})
    \label{eq11:target_distribution}
\end{equation}

\begin{table*}[t!]
    \vspace{+0.2cm}
    \centering
    \resizebox{\linewidth}{!}{
    \begin{tabular}{c c c c c c c c c c c c c c c c c c}
        \toprule        
        \multirow[c]{3}{*}{\textsc{Level}} & \multirow[c]{3}{*}{\textsc{Metrics}} & \multicolumn{4}{c}{\textsc{DEM}} & \multicolumn{4}{c}{\textsc{PUTN}} & \multicolumn{4}{c}{$\mathcal{M}_{\tau}$\textsc{-Only}} & \multicolumn{4}{c}{\textsc{TAO-Plan}} \\
        \cmidrule(lr){3-6} \cmidrule(lr){7-10} \cmidrule(lr){11-14} \cmidrule(lr){15-18}
        & & \multicolumn{2}{c}{\textsc{Vanilla}} & \multicolumn{2}{c}{\textsc{TAO-Ctr}} & \multicolumn{2}{c}{\textsc{PUTN-Ctr}} & \multicolumn{2}{c}{\textsc{TAO-Ctr}} & \multicolumn{2}{c}{\textsc{Vanilla}} & \multicolumn{2}{c}{\textsc{TAO-Ctr}} & \multicolumn{2}{c}{\textsc{Vanilla}} & \multicolumn{2}{c}{\textsc{TAO-Ctr}}\\
        \cmidrule(lr){3-4} \cmidrule(lr){5-6} \cmidrule(lr){7-8} \cmidrule(lr){9-10} \cmidrule(lr){11-12} \cmidrule(lr){13-14} \cmidrule(lr){15-16} \cmidrule(lr){17-18}
        & & \textsc{MPC} & \textsc{MPPI} & \textsc{MPC} & \textsc{MPPI} & \textsc{MPC} & \textsc{MPPI} & \textsc{MPC} & \textsc{MPPI} & \textsc{MPC} & \textsc{MPPI} & \textsc{MPC} & \textsc{MPPI} & \textsc{MPC} & \textsc{MPPI} & \textsc{MPC} & \textsc{MPPI} \\
        \midrule
        \multirow{3}{*}{\makecell{\textsc{Easy}}} 
            & \textsc{Success}  (\%) & 52.22 & 49.51 & 52.84 & 47.53 & 91.33 & 95.60 & 91.60 & 95.87 & 95.80 & 92.10 & 95.56 & 94.32 & 96.67 & 97.04 & \textbf{98.15}    & \underline{97.78} \\
            & \textsc{Step}     (k)  &  1.73 &  1.84 &  1.70 &  1.84 &  0.68 &  0.66 &  0.71 &  0.65 &  0.66 &  0.70 &  0.66 &  0.70 &  0.65 &  0.74 &  \textbf{0.60} &  \underline{0.63} \\
            & \textsc{Progress} (\%) & 71.16 & 71.34 & 71.25 & 69.69 & 93.59 & 97.03 & 94.21 & 97.21 & 96.96 & 97.56 & 96.75 & 97.56 & 98.06 & 98.34 & \textbf{98.76}    & \underline{98.68} \\
        \midrule
        \multirow{3}{*}{\makecell{\textsc{Plain}}} 
            & \textsc{Success}  (\%) & 22.96 & 22.96 & 19.51 & 23.83 & 83.47 & 89.33 & 76.80 & \underline{90.93} & 85.19 & 85.43 & 85.31 & 90.12 & 85.68 & 89.14 & 86.67 & \textbf{93.83} \\
            & \textsc{Step}     (k)  &  2.47 &  2.47 &  2.54 &  2.42 &  1.02 &  1.06 &  0.99 &  0.90 &  1.00 &  1.08 &  0.98 &  \underline{0.89} &  1.01 &  1.02 &  0.98 & \textbf{ 0.82} \\
            & \textsc{Progress} (\%) & 53.65 & 53.20 & 50.38 & 53.93 & 89.80 & 93.41 & 87.27 & 93.50 & 92.26 & 92.55 & 92.91 & 93.73 & 92.20 & \underline{94.00} & 93.16 & \textbf{96.79} \\
        \midrule
        \multirow{3}{*}{\makecell{\textsc{Hard}}} 
            & \textsc{Success}  (\%) &  6.79 &  7.16 &  6.67 & 12.96 & 40.27 & 55.33 & 50.00 & 61.20 & 38.52 & 50.86 & 40.62 & 62.10 & 61.11 & \underline{65.80} & 61.23 & \textbf{81.73} \\
            & \textsc{Step}     (k)  &  2.84 &  2.84 &  2.85 &  2.69 &  2.21 &  1.94 &  1.89 &  1.62 &  2.15 &  1.96 &  2.09 &  \underline{1.60} &  1.68 &  1.75 &  1.68 & \textbf{ 1.27} \\
            & \textsc{Progress} (\%) & 42.04 & 42.46 & 41.04 & 45.68 & 70.05 & 79.26 & 76.54 & 82.88 & 69.57 & 76.31 & 70.95 & 82.94 & 80.38 & \underline{83.74} & 81.25 & \textbf{89.50} \\
        \bottomrule
    \end{tabular}
    }
    \caption{\textbf{Comparison for Navigation Methods.} TAO (TAO-Plan + TAO-Ctr) outperforms other methods in every metric, especially at the hard level. The best results in each category are in \textbf{bold}, with the second best results \underline{underlined}.
    \label{tab1:navigation_results}
    \vspace{-0.5cm}
}
\end{table*}

The rejection sampling algorithm is presented in Alg. \ref{algo1:rejection_sampling}, where $\rho_{i}$ is the target distribution. The input $B = [x_{min}, x_{max}] \times [y_{min}, y_{max}]$ represents the boundary of the area perceived by the robot. In line 1, we use $B$ to generate proposal samples using the UniformSampling function. In line 5, $U(0,1)$ represents uniform distribution over the range $[0,1]$. Even though new nodes are not directly placed in the constrained region, the tree expansion process may create new edges that cross this area. To avoid this, we connect $N_{i}$ and $N_{j}$ with a path segment $\mathbf{x}_{i:j} = \{\mathbf{x}_{i}, \cdots, \mathbf{x}_{k}, \cdots, \mathbf{x}_{j}\}$ of length $l_{ij}$ composed of infinitesimal lengths $\Delta L$. Then, $\rho_{k}$ is evaluated at a point $\mathbf{x}_{k} \in \mathbf{x}_{i:j}$ using Eq. \ref{eq11:target_distribution}. Based on the notation above, we design the cost function $g$ over an edge $\mathcal{E}_{ij}=(\mathcal{N}_{i}, \mathcal{N}_{j})$ that connects two nodes as follows:

\begin{equation}
    g(\mathcal{E}_{ij}) = \Bigl(1 + \kappa \Bigl( \frac{1}{\sum_{k=i}^{j} \rho_{k}} - \vert \mathbf{x}_{i:j} \vert \Bigr) \Bigr) l_{ij},
    \label{eq12:cost_function}
\end{equation}

where $\kappa$ is a penalty weight for the proposed criterion, and $\vert \mathbf{x}_{i:j} \vert$ is denoted as the number of points in the path segment. In addition, we use a Reed-Shepp \cite{reeds1990optimal} curve to generate path segments to reduce the high degree of control freedom in mountainous terrain.
\newline

\subsubsection{Adaptive Control Optimization}
\label{adaptive_control_optimization}

We formulate a nonlinear model predictive controller (NMPC) that prevents getting stuck or damaged when following the feasible path generated by the planner at every moment. We use a skid-steering model \cite{kozlowski2004modeling} $f$ to predict the robot's state $\mathbf{s}=(x,y,\theta)$, which is as follows:

\begin{equation}
    \begin{bmatrix} x_{t+1} \\ y_{t+1} \\ \theta_{t+1} \end{bmatrix} = 
    \begin{bmatrix} x_{t} \\ y_{t} \\ \theta_{t} \end{bmatrix} + 
    \begin{bmatrix} \cos\theta_{t} & 0 \\ \sin\theta_{t} & 0 \\ 0 & 1 \end{bmatrix}
    \mathbf{u}_t \Delta t,
    \label{eq13:motion_model}
\end{equation}

where $\mathbf{u_t} = (\zeta_{t}, \omega_{t})$ is a control variable, denoting linear velocity and angular velocity, respectively, $\theta_{t}$ is heading angle and $\Delta t > 0$ is time interval. We utilize both apparent and relative traversability to optimize control parameters, as outlined below:

\vspace{-0.1cm}
\begin{equation}
    \begin{gathered}
        \min_{\mathbf{s}, \mathbf{u}} \sum_{k=t}^{t+T-1} \{ \|\mathbf{s}_k - \mathbf{s}_{k}^{d} \|^2_{Q} + \|\mathbf{u}_{k}\|^2_{R_{k}} + \|\mathbf{v}_k - \mathbf{v}_{k}^{d} \|^2_{W_{k}}\}\\
        + \lambda C_{\tau}(\mathbf{x}_{k:T}), \; s.t. \; \mathbf{s}_{k+1} = f(\mathbf{s}_{k}, \mathbf{u}_{k}) 
    \end{gathered}    
    \label{eq14:control_objective}
\end{equation}

\begin{equation} 
    R_{k} = \biggl(1 - \frac{1}{|\mathcal{K}(\mathbf{x}_k)|} \sum_{\mathbf{x}_j \in \mathcal{K}(\mathbf{x}_k)} \tau_{j} \biggl)^{-2} R
    \label{eq15:control_manitude_parameter}
\end{equation}

\begin{equation} 
    W_{k} = \bigl(1 + k_q \max(0, \tan \psi_{k})\bigr) W^{v},
    \label{eq16:velocity_error_parameter}
\end{equation}

 where $\mathbf{s}^{d}$ is desired state, $\mathbf{v}^{d}$ is velocity on $\mathcal{P}^{*}$, and $T$ represents prediction horizon. $Q \in \mathbb{R}^{3 \times 3}$ and $R \in \mathbb{R}^{2 \times 2}$ are positive definite matrices, where $Q = \mathrm{diag}(Q^x, Q^y, Q^\phi)$ and $R = \mathrm{diag}(R^{\zeta}, R^{\omega})$. In Eq. \ref{eq14:control_objective}, the first term minimizes the error between the predicted state and the desired state. The second term slows down the robot by adjusting the control variable magnitude based on traversability, as shown Eq. \ref{eq15:control_manitude_parameter}. The third term dynamically adjusts the robot's velocity by regulating $W_{k}$ through Eq. \ref{eq16:velocity_error_parameter}, which enables the robot to maneuver bumps or ditches. The last term considers the open space based on apparent traversability around predicted states. We sample points uniformly around the predicted states, then compute $C_{\tau}$ by averaging traversability for these random points, where $\lambda$ is a constant.

\section{Experiments} \label{experiment}

We designed our experiments to answer three questions:

\textbf{Q1.} How does our planning method compare to prior methods for generating a feasible path in extreme mountainous terrains?

\textbf{Q2.} Can our navigation method be generalized to rough and undulating terrains?

\textbf{Q3.} What are the benefits achieved through coupling resilient behavior to existing control methods?


\subsection{Simulation Settings}
We set a virtual environment with a mesh model of the real-world terrain on the PyBullet physics engine \cite{coumans2016pybullet}. The real-world terrain data is acquired as a point cloud via Clearpath Husky UGV mounted with LiDAR and transformed into the mesh. We collected 27 types of various terrains on campus and mountains, which have an average size of $22m \times 16m$. To demonstrate the robustness of TAO on rough and undulating terrains, we augment the terrain from 27 to 108 types by introducing four variations in the simulation. The variations are produced by adjusting the overall inclination in $5^{\circ}$ increments between $0^{\circ}$ and $15^{\circ}$. For the evaluation, we sort terrains based on the traversability score into levels of easy, plain, and hard. A higher level indicates rougher terrain with steeper inclinations. Tab. \ref{table:hyperparameter} describes the hyperparameters used for our experiments.

\begin{table*}[ht]
    \vspace{+0.2cm}
    \centering
    \resizebox{\linewidth}{!}{%
        \begin{tabular}{c c c c c c c c c c c c c c c c c}
        \toprule
        \multirow{2}{*}{\textsc{Level}} & \multicolumn{4}{c}{\textsc{AEG} (mm)} & \multicolumn{4}{c}{\textsc{Traversability} (\%)} & \multicolumn{4}{c}{\textsc{CTE} (cm)} & \multicolumn{4}{c}{\textsc{Inconsistency} (cm)} \\
        \cmidrule(lr){2-5} \cmidrule(lr){6-9} \cmidrule(lr){10-13} \cmidrule(lr){14-17}
        & \textsc{DEM} & \textsc{PUTN} & $\mathcal{M}_{\tau}$\textsc{-Only} & \textsc{TAO-Plan} & \textsc{DEM} & \textsc{PUTN} & $\mathcal{M}_{\tau}$\textsc{-Only} & \textsc{TAO-Plan} & \textsc{DEM} & \textsc{PUTN} & $\mathcal{M}_{\tau}$\textsc{-Only} & \textsc{TAO-Plan} & \textsc{DEM} & \textsc{PUTN} & $\mathcal{M}_{\tau}$\textsc{-Only} & \textsc{TAO-Plan} \\
        \midrule
        \textsc{Easy}  &  6.53 & 10.77 & 4.92 & \textbf{4.79} & 10.99 & 4.42 & 1.21 & \textbf{1.08} & 16.88 & 16.24 & \textbf{12.56} & 12.75 & 3.46 & 3.91 & 2.74 & \textbf{2.74} \\
        \textsc{Plain} &  8.67 & 12.38 & 5.32 & \textbf{5.26} & 13.67 & 4.82 & 1.20 & \textbf{1.15} & 18.14 & 19.09 & 12.57 & \textbf{12.45} & 3.82 & 3.79 & \textbf{2.76} & 2.78 \\
        \textsc{Hard}  & 14.72 & 13.77 & 8.85 & \textbf{7.94} & 13.98 & 5.71 & 1.78 & \textbf{1.47} & 23.79 & 22.64 & \textbf{21.06} & 21.56 & 4.35 & 3.32 & 3.08 & \textbf{3.06} \\
        \bottomrule
        \end{tabular}
    }%
    \caption{\textbf{Comparison for Path Planning Methods.} {\color{blue}TAO} consistently outperforms other methods in terms of AEG and Traversability. Although the computed trajectories have a larger CTE, there is no degradation in inconsistency.}
    \vspace{-0.3cm}
    \label{tab2:planning_results}
\end{table*}

\begin{figure*}
    \centering
    \includegraphics[width=500pt]{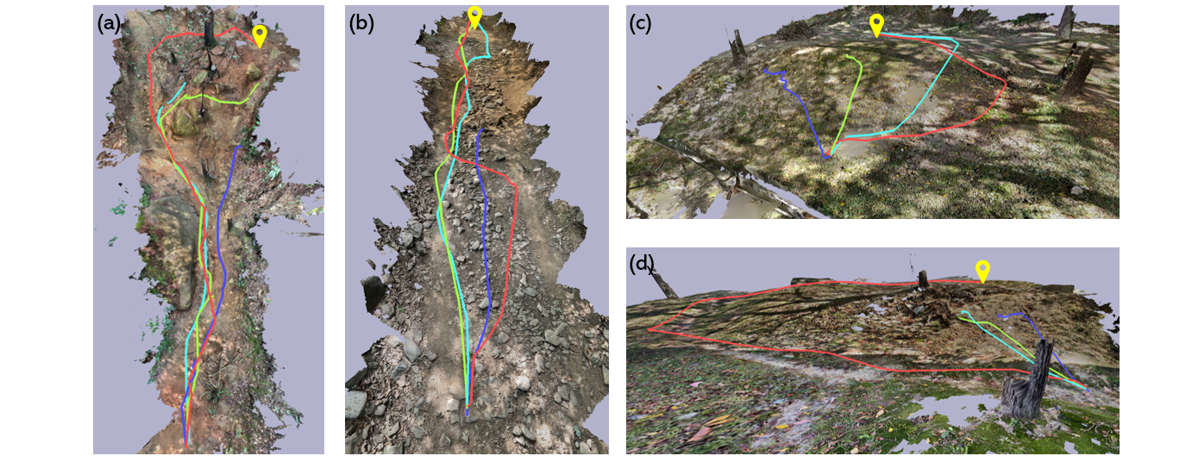}
    \caption{\textbf{Navigation in Mountainous Environments.} We compare 4 different path planning methods - DEM (purple), $\mathcal{M}_{\tau}$\textsc{-Only} (green), PUTN (cyan) and TAO-Plan (red) - using Adaptive MPPI controller. The goal point is indicated by a yellow marker. Our proposed method (TAO) successfully navigates in mountainous terrains by computing trajectories that circumvent bumpy and risky regions and dynamically adjust the robot's velocity.}
    \label{fig4:trajectories}
    \vspace{-0.5cm}
\end{figure*}

\subsection{Baselines and Evaluation Metrics}
We compare the effectiveness of our planning method (TAO-Plan) with three variations of the sampling-based planning method: DEM \cite{tanaka2015fuzzy}, PUTN \cite{jian2022putn}, and $\mathcal{M}_{\tau}$-Only. To ensure fairness in our comparison, all methods uniformly sample nodes for tree expansion. For the DEM, the cost is calculated based on a digital elevation map. 
In the case of PUTN, the methodology for evaluating terrain remains intact. Unlike $\mathcal{M}_{\tau}$-Only, PUTN utilizes the flatness feature instead of the bumpiness in Eq. \ref{eq4:bumpiness}. The $\mathcal{M}_{\tau}$-Only does not consider $\mathcal{M}_{\gamma}$ to calculate the cost, which serves as an ablation study for TAO-Plan. Moreover, we compare two control methods: Vanilla and TAO-Ctr. For Vanilla, we adopt MPC \cite{grune2017nonlinear} and MPPI \cite{williams2017model}, which are derivative-based and sampling-based approaches, respectively. For TAO-Ctr, we extend Vanilla method with adaptive control optimization, which demonstrates the ease of coupling our method with other controllers. Vanilla does not consider ground interaction, whereas TAO-Ctr considers the interaction by using the relative traversability $\Psi_{\tau}$. In the following contents, we denote TAO-Ctr as Adaptive for the ease of notion. In the case of PUTN, we employ the original control strategy from PUTN, referred to as PUTN-Ctr, which reduces velocity in regions with low traversability. Note that this approach solely relies on apparent traversability and does not incorporate relative traversability. We combine the aforementioned methods and evaluate the navigation performance. All evaluations are conducted with 30 trials for each terrain using the following metrics:

\begin{itemize}
\item \textbf{Average Elevation Gradient (AEG)} - The average of the elevation gradients experienced by the robot along the trajectory. The elevation gradient is calculated by the difference of the robot's elevation at any time instant.
\item \textbf{Traversability} - The summation of traversability of the nodes located along a planned trajectory. Note that the higher traversability value indicates that the ground is non-traversable. 
\item \textbf{Cross Tracking Error (CTE)} - The average distance error between the robot's position and the trajectory.
\item \textbf{Inconsistency} - The average distance error between the desired position and the actual position after executing a planned behavior.
\item \textbf{Success} - The rate at which the robot has successfully reached the goal while avoiding collisions or getting stuck. We assume that the robot is stuck when the difference in position and heading is insignificant for 30 seconds.
\item \textbf{Step} - The average number of steps the robot takes to reach the goal within a single episode. The maximum step is set to 3000.
\item \textbf{Progress} - The ratio of the distance the robot actually traveled to the distance of the planned path.
\end{itemize}

\subsection{Navigating Extreme Mountainous Terrains}

We show quantitative results in Tab. \ref{tab1:navigation_results}. TAO (TAO-Plan + TAO-Ctr) outperforms all the other methods by achieving successful navigation with the fewest number of steps. As shown in Progress, even in failure cases, TAO can travel the farthest distance before a collision or getting stuck. This is particularly evident at the hard level (\textbf{Q2}), where TAO outperforms the second-best baseline in success rate by a margin of +16\% (66\% to 82\%). As expected, the DEM performs well at the easy level but its performance sharply declines as the difficulty level increases. This demonstrates that height information alone is insufficient for a robot to navigate complex terrain. Both PUTN and $\mathcal{M}_{\tau}$-Only exhibit comparable performance to TAO in the easy and plain levels. However, in hard terrains, performance decreased as they failed to generate paths that circumvent bumpy and inclined regions. Fig. \ref{fig4:trajectories} visualize trajectories of four different path planning methods using an Adaptive MPPI controller. Fig. \ref{fig4:trajectories}(a) illustrates the example of hard-level terrain with the highest traversability score. The right path includes a narrow and steep slope below the goal point. In order to reach the goal, the robot must follow the left path. DEM fails in reaching the goal due to the presence of a higher sill, which is beyond the robot's ground clearance. PUTN encounters difficulty in identifying the tree roots, which leads to getting stuck and being unable to reach the goal. $\mathcal{M}_{\tau}$-Only computes a trajectory that circumvents the sill but still takes the right passage. TAO generates the left path and successfully navigates to the goal through resilient path planning and control.

\begin{figure}[t!]
    \vspace{+0.15cm}
    \centering
    \begin{subfigure}[b]{0.21\textwidth}
        \includegraphics[width=\textwidth]{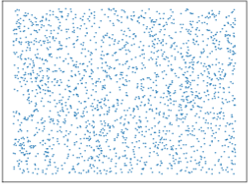}
        \vspace{-0.6cm}
        \captionsetup{justification=centering}
        \caption{Uniform Sampling}
        \label{fig5a:uniform_sampling}
    \end{subfigure}
    \begin{subfigure}[b]{0.21\textwidth}
        \includegraphics[width=\textwidth]{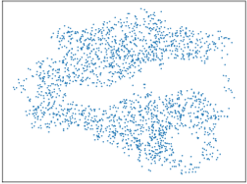}
        \vspace{-0.6cm}
        \captionsetup{justification=centering}
        \caption{Rejection Sampling}
        \label{fig5b:rejection_sampling}
    \end{subfigure}
    \begin{subfigure}[b]{0.21\textwidth}
        \includegraphics[width=\textwidth]{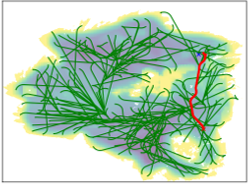}
        \vspace{-0.6cm}
        \captionsetup{justification=centering}
        \caption{{$\mathcal{M}_{\tau}$\textsc{-Only}}}
        \label{fig5c:uniform_path}
    \end{subfigure}
    \begin{subfigure}[b]{0.21\textwidth}
        \includegraphics[width=\textwidth]{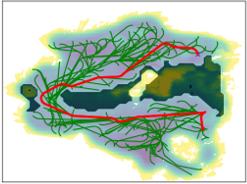}
        \vspace{-0.6cm}
        \captionsetup{justification=centering}
        \caption{\textsc{TAO-Plan}}
        \label{fig5d:rejection_path}
    \end{subfigure}
    \caption{\textbf{Path Generation Process.} The difference between (a) and (b) shows the sampling bias. (c) and (d) also visualizes the search tree and optimal path overlaid on the traversability map. The search tree is depicted as green solid lines, while the selected optimal path is shown as red solid lines.}
    \vspace{-0.3cm}
    \label{fig5:sampling}
\end{figure}

\begin{figure}[t!]
\centering
\begin{subfigure}{0.5\textwidth}
    \centering
    \includegraphics[width=0.65\columnwidth]{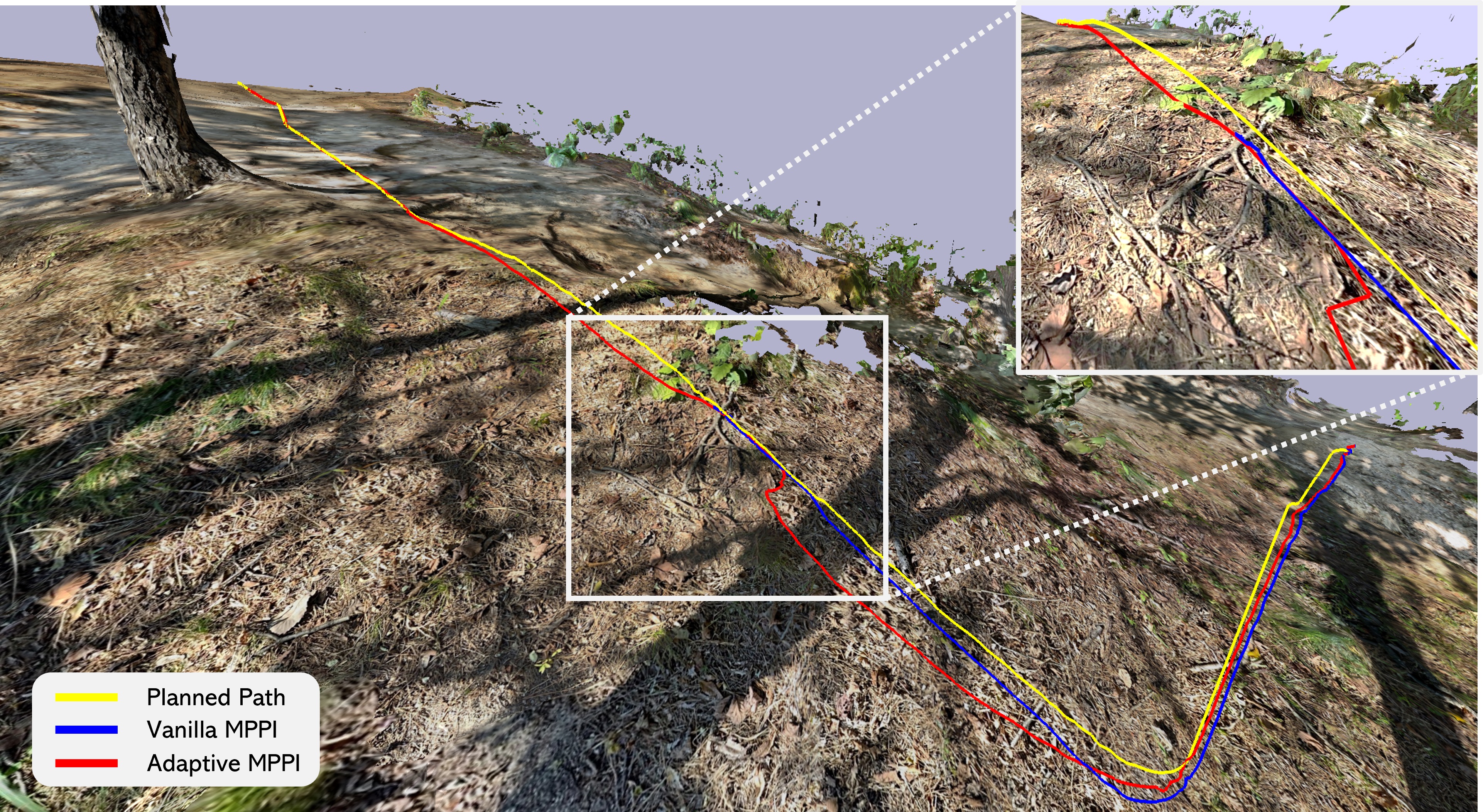}
    \captionsetup{justification=centering}
    \vspace{-0.1cm}
    \caption{Simulated Terrain}
    \label{fig6a:resilient_behavior_terrain}
\end{subfigure}
\newline
\begin{subfigure}[b]{0.47\columnwidth}
    \includegraphics[width=\columnwidth]{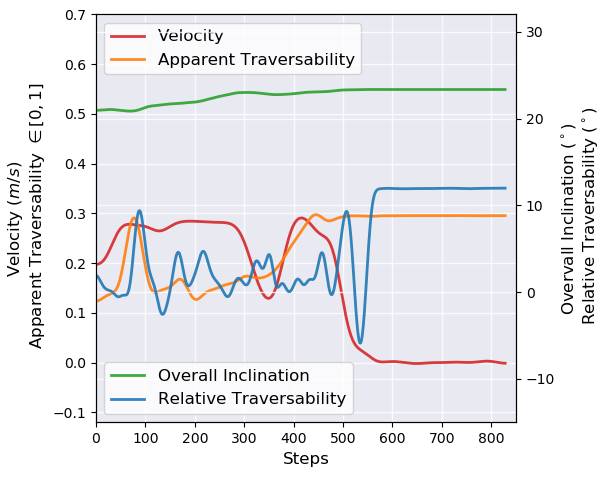}
    \captionsetup{justification=centering}
    \vspace{-0.5cm}
    \caption{\textsc{Vanilla} MPPI}
    \label{fig6b:vel_plot_without_resilient}
\end{subfigure}
\begin{subfigure}[b]{0.47\columnwidth}
    \includegraphics[width=\columnwidth]{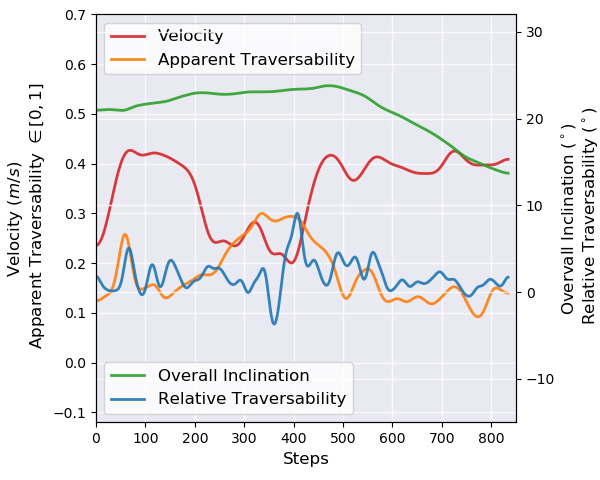}
    \captionsetup{justification=centering}
    \vspace{-0.5cm}
    \caption{\textsc{Adaptive} MPPI}
    \label{fig6c:vel_plot_with_resilient}
\end{subfigure}
\caption{\textbf{Resilient Behavior for Adaptive Control.} (a) shows the trajectories that the robot has followed using each control method. (b) Without resilient behavior, the robot's velocity becomes zero due to bumpy terrain with a steep inclination. (c) With resilient behavior, the robot dynamically adjusts the velocity according to the relative traversability. Note that \textsc{Adaptive} refers TAO-Ctr.}
\vspace{-0.6cm}
\label{fig6:vel_plot}
\end{figure}

\subsection{Traversability-aware Path Planning}

Fig. \ref{fig4:trajectories}(b) depicts a steeply slanting environment with scattered small and large rocks, which causes the robot to get stuck while moving across the region. DEM generates a trajectory that passes through the bumpy region, but since it lacks the ability to analyze intricate terrain characteristics beyond the height map, the robot eventually gets stuck. PUTN tends to stick close to or step over gravel in narrow passages. Although $\mathcal{M}_{\tau}$-Only avoids the bumpy middle region clearly by considering the bumpiness of terrain features in Eq. \ref{eq4:bumpiness}, it generates a trajectory toward a relatively narrow passageway on the left. In contrast, our proposed method chooses the wider right passageway and successfully navigates through the least undulating region to reach the goal (\textbf{Q1}).

In Fig. \ref{fig4:trajectories}(c-d), there are steep hills between the starting and the goal point with rough ground. 
DEM and $\mathcal{M}_{\tau}$-Only focus on the shortest distance, they fail to reach the goal due to slippage or roll-over. PUTN can reach the goal in Fig. \ref{fig4:trajectories}(c), but shows an unstable trajectory passing over a high slope. In Fig. \ref{fig4:trajectories}(d), similar to prior methods, it fails to generate a trajectory to circumvent the risky regions. On the other hand, our approach designates the rough area as a constraint and computes a trajectory that bypasses it, allowing the robot to successfully reach the goal. The ability of TAO-Plan to generate such a path is attributed to the proposed rejection sampling in Eq. \ref{eq11:target_distribution} and cost function in Eq. \ref{eq12:cost_function}. The path generation process in Fig. \ref{fig4:trajectories}(d) is depicted in Fig. \ref{fig5:sampling}. 

Tab. \ref{tab2:planning_results} shows performance comparison in terms of the path planning method. All methods used an Adaptive MPPI controller. In most cases, the TAO-Plan outperforms other methods, especially with traversability and AEG scores. This result implies that apparent traversability induces generating a stable path without large fluctuations. Meanwhile, even if CTE increases, there is no significant difference in inconsistency.

\subsection{Resilient Behavior for Control}

We analyze the effect of adaptive control optimization (Section \ref{adaptive_control_optimization}). We use the same planned path (TAO-Plan) to compare control strategies fairly. As shown in Tab. \ref{tab1:navigation_results}, applying TAO-Ctr (Adaptive) leads to performance improvement in most methods. In particular, the performance is significantly improved at the hard level due to the strategy of regulating the velocity considering relative traversability, slowing down in dangerous areas, and accelerating in areas where the robot is likely to get stuck (\textbf{Q3}). Also, TAO-Ctr shows better performance compared to PUTN-Ctr in hard terrains within the PUTN planner, highlighting that the approach of regulating velocity based on the robot's inertial state is more effective than simply decreasing velocity based on solely geometric representation.

Fig. \ref{fig6:vel_plot} illustrates how our method dynamically adjusts the robot's velocity in an environment that has a steep inclination and uneven terrain. To pass through such an environment, the robot must momentarily increase its velocity. In Vanilla MPPI, the robot gets stuck in areas where the relative traversability suddenly increases, resulting in the velocity eventually dropping to zero. In contrast, in Adaptive MPPI, the robot momentarily accelerates when relative traversability suddenly increases.

Adaptive control optimization can be easily integrated into existing controllers. We compare the navigation performance of different controllers, MPC and MPPI in Tab. \ref{tab1:navigation_results}. Using MPC shows better performance at the easy level. On the other hand, as the environment becomes more challenging, MPPI, which is less sensitive to the motion model, shows better performance.

\begin{figure}[t!]
    \centering
    \includegraphics[width=0.75\columnwidth]{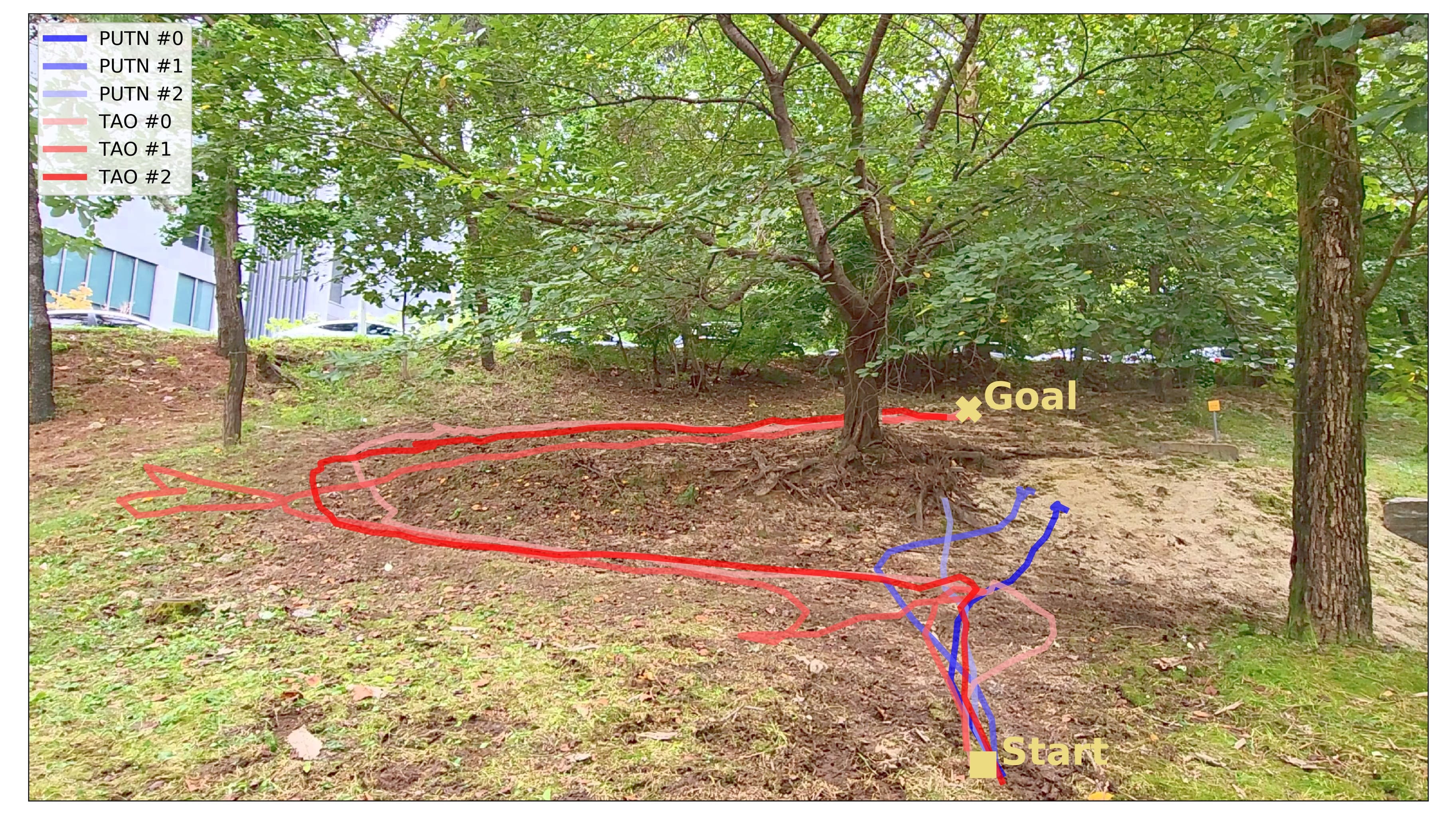}
    \caption{\textbf{Real-world Navigation.} TAO (TAO-plan + TAO-Ctr) successfully reaches the goal while PUTN fails.}
    \vspace{-0.5cm}
    \label{fig8:real_world_navigation}  
\end{figure}

\subsection{Real-world Navigation}
We set up a real Husky robot similar to the simulated robot. Intel NUC with i7 CPU and OS0-64 LiDAR is mounted on the robot. We utilize FAST-LIO2\cite{xu2022fast} for localization and RRT$\mathcal{^X}$\cite{otte2016rrtx} for generating a search tree.

We compare TAO with PUTN \cite{jian2022putn} in a real-world environment. The experiment is conducted in the same environment as Fig. \ref{fig4:trajectories}(d) and Fig. \ref{fig5:sampling} and the result is shown in Fig. \ref{fig8:real_world_navigation}. PUTN generates a path that cannot circumvent the steep slope, resembling the behavior of DEM. Moreover, the real-world environment exhibits more slippery than the simulated environment, PUTN fails to progress as expected due to a strategy of reducing velocity in high traversability areas. On the other hand, generates a circumvent path by marking a highly inclined area as a constraint and successfully reaches the goal by regulating the velocity. Note that TAO \#1 exhibited a substantial deviation from other trajectories as a result of slippage.

\section{Conclusion} \label{conclusion}

In this paper, we present a method for safely and efficiently navigating in extreme mountainous terrain. We assess the terrain with precisely designed features and generate a path that the robot can faithfully execute. Resilient behavior for robust navigation is achieved through adaptive control optimization by considering ground-robot interaction. Furthermore, our method can easily be combined with existing planning and control methods. While our method focuses solely on geometric features, incorporating visual features could be a valuable extension to enhance generalizability across seasonal changes in terrains.

\begin{table}[t!]        
    \vspace{+0.2cm}
    \begin{center}
        \begin{tabular}{c c}
        \toprule
        \textsc{Hyperparameter} & \textsc{Value} \\
        \midrule
        $T$, $N$, $H$, $W$, $\Delta L$, $\Delta t$ &  20, 2000, 500, 500, 0.1, 0.1 \\                        
        $\alpha_{1}$, $\alpha_{2}$, $\alpha_{3}$, $\kappa$, $k_{q}$ & 0.3, 0.3, 0.4, 5, 5 \\                   
        $r_{min}$, $r_{max}$, $\epsilon_{slop}$, $\epsilon_{spar}$, $\epsilon_{bump}$ & 0.2, 0.8, 0.7, 0.6, 0.5 \\                
        $Q^x$, $Q^y$, $Q^\phi$, $R^{\zeta}$, $R^{\omega}$, $W^v$ & 10, 10, 1, 0.5, 0.5, 3 \\                
        \bottomrule
        \end{tabular}
    \end{center}    
    \captionsetup{justification=centering}    
    \vspace{-0.2cm}
    \caption{\textbf{Hyperparameter.}}
    \label{table:hyperparameter}
    \vspace{-0.5cm}
\end{table} 



\ifCLASSOPTIONcaptionsoff
  \newpage
\fi
\bibliographystyle{IEEEtran}
\bibliography{IEEEabrv, ref}

\end{document}